\documentclass{ecai}  

\usepackage{nomencl}
\makenomenclature
\usepackage{etoolbox}
\renewcommand\nomgroup[1]{%
  \item[\bfseries
  \ifstrequal{#1}{D}{Data}{%
  \ifstrequal{#1}{M}{Module}{%
  \ifstrequal{#1}{P}{Hyper-parameter}{%
  \ifstrequal{#1}{L}{Loss Function}{}}}}%
]}

\usepackage{graphicx}
\usepackage{latexsym}
\usepackage{hyperref}
\usepackage{latexsym}
\usepackage{amssymb}
\usepackage{amsmath}
\usepackage{amsthm}
\usepackage{booktabs}
\usepackage{enumitem}
\usepackage{graphicx}
\usepackage{color}
\usepackage{algorithm}
\usepackage{algorithmic}
\usepackage{color} 
\usepackage{framed}
\usepackage[table,xcdraw]{xcolor}
\usepackage{multirow}
\definecolor{deepgreen}{rgb}{0, 0.6, 0.2}
\definecolor{githubRed}{rgb}{0.753, 0.212, 0.298}

\hypersetup{
    colorlinks=true,
    linkcolor=blue,     
    filecolor=magenta,  
    urlcolor=githubRed      
}

\begin{document}

\begin{frontmatter}

\title{MakeupAttack: Feature Space Black-box Backdoor Attack on Face Recognition via Makeup Transfer}

\author[A,B]{\fnms{Ming}~\snm{Sun}}
\author[A,B]{\fnms{Lihua}~\snm{Jing}\thanks{Corresponding Author. Email: jinglihua@iie.ac.cn}}
\author[A,B]{\fnms{Zixuan}~\snm{Zhu}}
\author[A,B]{\fnms{Rui}~\snm{Wang}} 

\address[A]{Institute of Information Engineering, Chinese Academy of Sciences, Beijing, China}
\address[B]{School of Cyber Security, University of Chinese Academy of Sciences, Beijing, China}

\begin{abstract}
Backdoor attacks pose a significant threat to the training process of deep neural networks (DNNs). As a widely-used DNN-based application in real-world scenarios, face recognition systems once implanted into the backdoor, may cause serious consequences. Backdoor research on face recognition is still in its early stages, and the existing backdoor triggers are relatively simple and visible. Furthermore, due to the \textit{perceptibility}, \textit{diversity}, and \textit{similarity} of facial datasets, many state-of-the-art backdoor attacks lose effectiveness on face recognition tasks. In this work, we propose a novel feature space backdoor attack against face recognition via makeup transfer, dubbed MakeupAttack. In contrast to many feature space attacks that demand full access to target models, our method only requires model queries, adhering to black-box attack principles. In our attack, we design an iterative training paradigm to learn the subtle features of the proposed makeup-style trigger. Additionally, MakeupAttack promotes trigger diversity using the adaptive selection method, dispersing the feature distribution of malicious samples to bypass existing defense methods. Extensive experiments were conducted on two widely-used facial datasets targeting multiple models. The results demonstrate that our proposed attack method can bypass existing state-of-the-art defenses while maintaining \textit{effectiveness}, \textit{robustness}, \textit{naturalness}, and \textit{stealthiness}, without compromising model performance. Our code is available at \href{https://github.com/AaronSun2000/MakeupAttack}{https://github.com/AaronSun2000/MakeupAttack}.
\end{abstract}

\end{frontmatter}

\begin{figure*}[ht]
    \centering
    \includegraphics[width=1.9\columnwidth]{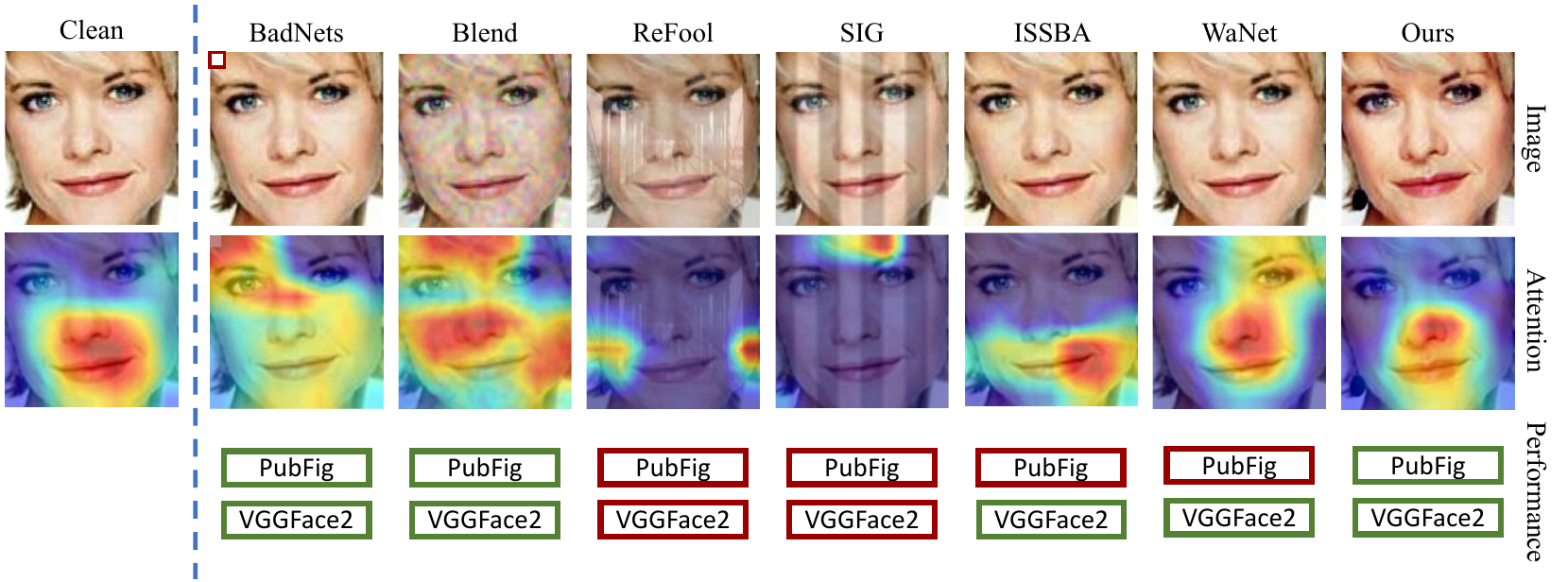}
    \caption{Comparison with existing backdoor attack methods. \textbf{TOP:}  the benign sample and different malicious samples generated by BadNets, Blend, ReFool, SIG, ISSBA, WaNet, and our method (MakeupAttack); \textbf{Middle:} attention maps generated by Grad-Cam; \textbf{Bottom:} the red box represents the dataset where the attack \textcolor{red}{fails}, while the green box represents the dataset where the attack \textcolor{deepgreen}{succeeds}.}
    \label{fig: intro}
\end{figure*}

\section{Introduction}

Deep neural networks (DNNs) have been widely deployed in many real-world visual scenarios, such as autonomous driving~\cite{grigorescu2020survey,wen2022deep}, intelligent medical diagnosis~\cite{litjens2017survey} and face recognition~\cite{wang2021deep}.  As model complexity explodes, training DNNs from scratch requires substantial resources and time. Therefore, third-party platforms or models are widely adopted, leading to hidden dangers, one of which is backdoor attacks.

Previous studies~\cite{li2022backdoor} show that DNNs are vulnerable to backdoor attacks during the training stage. Adversaries can easily implant potential backdoors into models by data poisoning. The attacked model will output predefined results when activated by the trigger on malicious samples while behaving normally on benign samples.

The face recognition (FR) system, a commonly employed DNN-based application, can pose security risks if attacked by backdoors, making it susceptible to exploitation by adversaries. However, few works have focused on the vulnerability of FR systems. The trigger patterns are conspicuous and the methods are relatively naive, including facial markings~\cite{xue2021backdoors}, accessories~\cite{wenger2020backdoor}, and image-blending~\cite{chen2017targeted}, which can be easily detected and mitigated by existing defenses. 

Many high-performance backdoor attacks are inevitably weakened in FR tasks, due to the limitations imposed by the characteristics of facial datasets:
(1) \textbf{Perceptibility.} The high perceptual acuity for facial features means that even subtle alterations to a face may be readily noticeable upon human inspection, posing a significant challenge to attack stealthiness.
(2) \textbf{Diversity.} Images of the same identity can exhibit variations in clarity, background, lighting conditions, posture, expression, and other aspects, resulting in excessive intra-class variance. Consequently, triggers embedded within the image may be overlooked by the model due to their overly small magnitude, leading to attack failures.
(3) \textbf{Similarity.} FR systems employ strict criteria to distinguish between individuals due to the inherent similarities in facial appearances. During backdoor training, data poisoning alters the model's decision-making process, diminishing its performance on benign samples.

In this paper, we propose MakeupAttack, a novel backdoor attack that utilizes makeup styles as the trigger pattern in the feature space. Unlike conspicuous markings~\cite{xue2021backdoors} or perturbations~\cite{gu2019badnets,chen2017targeted}, makeup triggers are more compatible with facial images, resulting in more natural-looking malicious samples. Figure~\ref{fig: intro} provides a comparison between our attack and existing methods across three key aspects: poisoned samples, model attention, and attack performance.

Given that the proposed trigger operates within the feature space, \textit{effectively enabling target models to learn subtle makeup-style triggers} poses a significant challenge. To address this concern, we introduce an iterative backdoor attack paradigm. Through mutual guidance between the trigger generator and the target model, the generator produces more potent malicious samples, thereby enhancing the attack effectiveness on the target model.  Unlike most feature space attacks~\cite{cheng2021deep,zhao2022defeat} with full access to target models, MakeupAttack adheres to a black-box attack setting, necessitating only model querying and data poisoning. Furthermore, we employ adaptive selection to promote trigger diversity. This entails malicious samples adaptively selecting appropriate reference images for makeup transfer, thus dispersing the feature distribution of malicious samples and circumventing many existing defenses.

To the best of our knowledge, MakeupAttack is the first attempt at employing configurable makeup styles as trigger patterns with a joint training framework in backdoor attacks. Our contributions can be summarized as follows:
(1) We propose MakeupAttack, a novel feature space backdoor attack via makeup transfer. This approach seamlessly combines \textit{effectiveness}, \textit{robustness}, \textit{naturalness}, and \textit{stealthiness}.
(2) We devise an iterative training paradigm for the trigger generator and the target model. This paradigm ensures that the target model comprehensively learns the subtle features of our triggers. To promote trigger diversity, we propose the adaptive reference image selection method.
(3) Extensive experiments across diverse facial datasets and network architectures validate the effectiveness, robustness, and resilience of our methods against various defenses.
(4) We construct high-quality malicious datasets to facilitate future research in this domain.


\section{Related Work}

\subsection{Poisoning-based Backdoor Attack}

BadNets~\cite{gu2019badnets} is the first backdoor attack on DNNs using a static patch as the trigger.  Subsequently, several attacks~\cite{liu2018trojaning,nguyen2020input} emerge, employing predefined patches or watermarks as triggers. However, these static patches or watermarks are easily detectable due to their conspicuous nature. In response, researchers have sought stealthier backdoor attack methods. ReFool~\cite{liu2020reflection}  exploits physical reflection to improve trigger naturalness. WaNet~\cite{nguyen2020wanet} adopts image warping as a distinctive trigger pattern. ISSBA~\cite{li2021invisible} utilizes image steganography to generate the invisible, sample-specific triggers. 

Except for pixel-level backdoor attacks, feature space attacks have also gained increasing attention from researchers. DFST~\cite{cheng2021deep} leverages CycleGAN to generate style-transferred poisoned samples. DEFEAT~\cite{zhao2022defeat} employs adaptive imperceptible perturbation as triggers and constrains latent representation during backdoor training to enhance resistance to defenses. Despite offering superior stealthiness and defense resilience,  many feature space attacks require full access to the training process,  limiting their applicability in real-world scenarios. In contrast, our approach not only generates natural and stealthy triggers in the feature space but is also compatible with black-box settings. 

Backdoor attack methods targeting face recognition remain relatively basic. Among them, the most prevalent approach involves facial accessories~\cite{chen2017targeted,wenger2020backdoor} or image-blending techniques~\cite{chen2017targeted}. Additionally, BHF2~\cite{xue2021backdoors} leverages specially-designed marks on eyebrows or beard as triggers. FaceHack~\cite{sarkar2021facehack} attempts to utilize off-the-shelf filters or APIs for directional correction of facial features, expressions, or age,  yet it fails to achieve significant attack effectiveness.  Our method surpasses these existing approaches in terms of both naturalness and effectiveness.

\subsection{Backdoor Defense}

Various defense strategies exist for mitigating backdoor attacks. Some existing studies leverage specific characteristics to detect malicious samples. STRIP~\cite{gao2019strip} discovers that sample superimposition has a relatively minor impact on model predictions of poisoned samples. Februus~\cite{doan2020februus} utilizes GradCAM~\cite{selvaraju2017grad} to identify potential triggers. Signature Spectral~\cite{tran2018spectral} demonstrates that backdoor attacks often leave discernible traces in the spectrum of the covariance of feature representation. Another method focuses on removing backdoors from poisoned models. Fine-Pruning~\cite{liu2018fine} identifies differences in activation value on malicious samples to screen out compromised neurons. NAD~\cite{li2020neural} employs a teacher network to guide the fine-tuning of a backdoored student network on a small set of benign samples. CLP~\cite{zheng2022data} employs channel Lipschitz constants to prune channels and repair backdoored models. A third category of methods diagnoses models using reversed triggers. Neural Cleanse~\cite{wang2019neural} is the first trigger synthesis-based defense, utilizing anomaly detection to identify the target label and corresponding trigger pattern. Subsequently, similar methods like  ABS~\cite{liu2019abs}, and DeepInspect~\cite{chen2019deepinspect} have emerged. Most existing defenses rely on the assumption of latent separability between benign and malicious samples, which was challenged by our method.  

\subsection{Makeup Transfer}

Makeup transfer, a technique employed to adapt facial images to specific makeup styles, has gained widespread adoption in the industry. BeautyGAN~\cite{li2018beautygan} introduces an end-to-end network based on a dual-input GAN to facilitate both makeup transfer and removal simultaneously. LADN~\cite{gu2019ladn} employs multiple overlapping local discriminators to achieve more precise transfer for makeup details. PSGAN~\cite{Jiang_2020_CVPR} addresses the challenge of transferring makeup across large poses and expression differences, enabling partial and interpolated makeup transfer. We incorporate an advanced makeup transfer framework into our backdoor attack paradigm, enhancing the naturalness and stealthiness of the transfer effect while also equipping it with backdoor attack capabilities.


\begin{figure*}[ht]
    \centering
    \includegraphics[width=2\columnwidth]{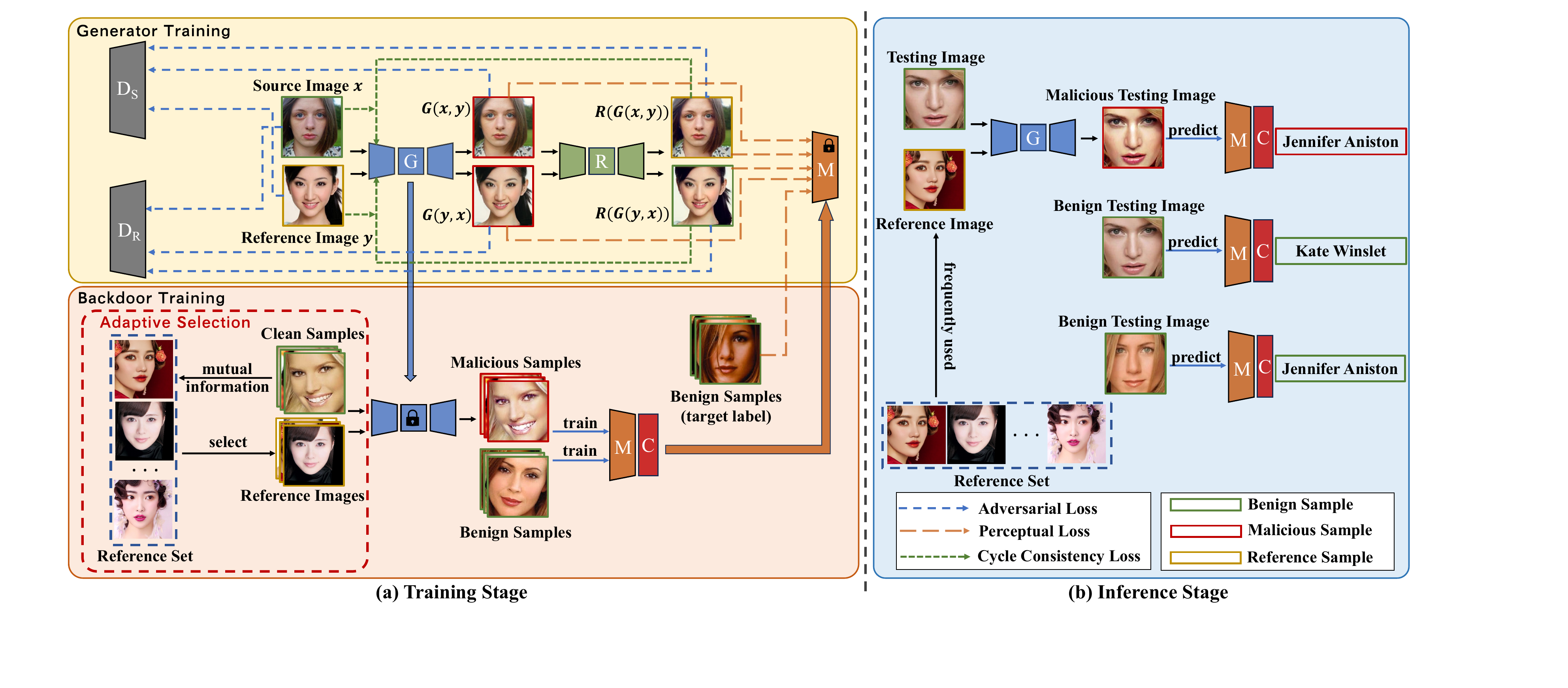}
    \caption{Overview of MakeupAttack. In the training stage, target models and trigger generators train alternatively, mutually guiding each other. Generator training and poisoned data updating proceed concurrently in the background, without disrupting the training procedure of target models. In the inference stage, the target model misclassifies malicious samples as the target label, while behaving normally on benign samples.}
    \label{fig2}
\end{figure*}

\section{Threat Model}

\subsection{Adversary's Capacities}

MakeupAttack follows the black-box attack settings. In the training stage, adversaries can only query the target models and poison part of the training data. In the inference stage, adversaries are not permitted to manipulate inference components. This threat model is particularly suitable for scenarios involving third-party platforms or APIs.

\subsection{Adversary's Goals}
\label{3.2}

\noindent \textbf{Effectiveness.} Target models should achieve a high attack success rate while maintaining performance on benign samples.

\noindent \textbf{Naturalness.} The trigger should be natural and imperceptible to both human visual perception and detection systems.

\noindent \textbf{Stealthiness.} Poisoned samples should exhibit subtle modifications, with a low poisoning rate to evade detection.

\noindent \textbf{Robustness.} The attack methods should demonstrate effectiveness across diverse datasets with varying scales and qualities, as well as multiple target models with different network structures.

\noindent \textbf{Resistance to Defenses.} The attack should be capable of bypassing a range of defense mechanisms.


\section{Method}

In this section, we first outline the MakeupAttack pipeline and then elaborate on each module individually. Figure \ref{fig2} demonstrates the overview of our method. 

\subsection{Overview}

During the training stage, the generator training phase and backdoor training phase iterate and mutually guide each other to facilitate more effective backdoor implantation into target models. In the generator training phase, we train the trigger generator using a  PSGAN-based framework, supplemented with a rectification module $R$ to ensure cycle consistency. In the backdoor training phase, we first construct a reference image set to specify multiple makeup styles. We then utilize the pre-trained generator to generate malicious samples and conduct the training procedure using both benign and malicious samples. After epochs, adversaries retrieve the currently saved optimal target model and guide the generator to undergo fine-tuning. In this fine-tuning phase, perception loss related to the target model is introduced into the original framework to guide the generator in creating more potent malicious samples. Subsequently, adversaries utilize the fine-tuned generator to regenerate malicious samples and update the corresponding dataset.

During the test stage, we expect the backdoored model to accurately predict benign samples while misclassifying the malicious samples produced by our generator as the predefined identity.

For each training sample, we employ mutual information to select the most suitable reference image from the reference set; while for test samples, we use the most frequently used reference image for transfer. This approach disperses the features of malicious samples, attenuates the distinct boundary with benign samples, and effectively bypasses many detection-based defenses.

\subsection{Generator Pre-training}
\label{4.2}
We denote the source domain and the reference domain as $\mathbf{S}$ and $\mathbf{R}$, respectively. Let $\boldsymbol{s}$ represent a source image sampled from $\mathbf{S}$ and $\boldsymbol{r}$ represent a reference image sampled from $\mathbf{R}$. During the generator training phase, we train the trigger generator $G$ to produce the transferred image $\tilde{\boldsymbol{s}}=G(\boldsymbol{s}, \boldsymbol{r})$. The transferred image retains the identity information of the source image $\boldsymbol{s}$ and the makeup style of the reference image $\boldsymbol{r}$, while also processing the potential for backdoor poisoning.

To achieve this, we employ a PSGAN-based framework to train the trigger generator $G$ for makeup transfer. We utilize two discriminators $D_S$ and $D_R$ for the source domain and the reference domain to enhance the authenticity of generated images. Additionally, a rectification module $R$ is integrated to ensure cycle consistency.

\textbf{Rectification Module and Cycle Consistency Loss.} Given that the generator $G$ is tasked with both makeup transfer and data poisoning, maintaining cycle consistency based solely on the original framework poses challenges. We hypothesize that the generated samples $G(\boldsymbol{s},\boldsymbol{r})$ do not directly transfer to the reference domain $\mathbf{R}$, but rather shift to what we term a malicious domain $\mathbf{R^M}$. Consequently, the recovered sample $G(G(\boldsymbol{s},\boldsymbol{r}),\boldsymbol{s})$ may fail to transition back to the source domain $\mathrm{S}$. To address this, we utilize a rectification module $R$ to correct the domain offset problem, thereby ensuring cycle consistency. Specifically, we employ a residual-in-residual dense block (RRDB)~\cite{wang2018esrgan} as the rectification module $R$, and reconstruct the domain transfer loop, i.e. $\mathbf{S}\rightarrow \mathbf{R^M}\rightarrow \mathbf{R}\rightarrow \mathbf{S^M}\rightarrow \mathbf{S}$. The rectified cycle consistency loss $L^{cyc}$ can be formulated as follows:
\begin{equation}
\begin{split}
    L_{G}^{cyc}=\mathrm{E}_{(\boldsymbol{s},\boldsymbol{r})}[||R(G(R(G(\boldsymbol{s},\boldsymbol{r})),\boldsymbol{s}))-\boldsymbol{s}||_1] \\
    +\mathrm{E}_{(\boldsymbol{s},\boldsymbol{r}) }[||R(G(R(G(\boldsymbol{r},\boldsymbol{s})),\boldsymbol{r}))-\boldsymbol{r}||_1].
\end{split}
\end{equation}

\textbf{Adversarial Loss.} We adopt adversarial loss $L^{adv}$ to guide the training of the trigger generator $G$ and two domain discriminators $D_S$, $D_R$, which can be formulated as follows:
\begin{equation}
\begin{split}
    L^{adv}_{D_{S}}=\mathrm{E}_{(\boldsymbol{s},\boldsymbol{r})}[-\log{D_S(\boldsymbol{s})}-\log{(1-D_S(G(\boldsymbol{r},\boldsymbol{s})))}],\\
    L^{adv}_{D_{R}}=\mathrm{E}_{(\boldsymbol{s},\boldsymbol{r})}[-\log{D_R(\boldsymbol{r})}-\log{(1-D_R(G(\boldsymbol{s},\boldsymbol{r})))}],
\end{split}
\end{equation}
\begin{equation}
    L^{adv}_{G}=\mathrm{E}_{(\boldsymbol{s},\boldsymbol{r}) }[-\log{D_S(G(\boldsymbol{r},\boldsymbol{s}))}-\log{D_R(G(\boldsymbol{s},\boldsymbol{r}))}]
\end{equation}

The adversarial loss also guides the rectification module through discriminators, which can be formulated as follows:
\begin{equation}
\begin{split}
    L^{adv}_{R}=\mathrm{E}_{(\boldsymbol{s},\boldsymbol{r})}[-\log{D_S(R(G(\boldsymbol{r},\boldsymbol{s})))}]\\
    +\mathrm{E}_{(\boldsymbol{s},\boldsymbol{r})}[-\log{D_R(R(G(\boldsymbol{s},\boldsymbol{r})))}].
\end{split}
\end{equation}

\textbf{Makeup Loss.} We introduce makeup loss~\cite{li2018beautygan} to provide coarse guidance for makeup transfer. Specifically, we first parse masks for lips, skin, and eye shadow. Then, we apply histogram matching on these regions and combine them into a pseudo-ground-truth $HM(\boldsymbol{s},\boldsymbol{r})$. The makeup loss is formulated as follows:
\begin{equation}
\begin{split}
    L_G^{mk} = \mathrm{E}_{(\boldsymbol{s},\boldsymbol{r}) }[||G(\boldsymbol{s},\boldsymbol{r})-HM(\boldsymbol{s},\boldsymbol{r})||_2]\\
    +\mathrm{E}_{(\boldsymbol{s},\boldsymbol{r}) }[||G(\boldsymbol{r},\boldsymbol{s})-HM(\boldsymbol{r},\boldsymbol{s})||_2],
\end{split}
\end{equation}
\begin{equation}
\begin{split}
    L_R^{mk} = \mathrm{E}_{(\boldsymbol{s},\boldsymbol{r})}[||R(G(\boldsymbol{s},\boldsymbol{r}))-HM(\boldsymbol{s},\boldsymbol{r})||_2]\\
    +\mathrm{E}_{(\boldsymbol{s},\boldsymbol{r})}[||R(G(\boldsymbol{r},\boldsymbol{s}))-HM(\boldsymbol{r},\boldsymbol{s})||_2].
\end{split}
\end{equation}

\textbf{Regularization Loss.} To safeguard the key information from the source image $\boldsymbol{s}$ and control the magnitude of facial modification, we utilize $l_1$ norm and LPIPS\footnote{LPIPS  measures perceptual similarity between two images.} to constrain image generation. The regularization loss can be formulated as follows:
\begin{equation}
\begin{split}
    L_{G,R}^{reg}=\mathrm{E}_{\boldsymbol{s}}[||R(G(\boldsymbol{s},\boldsymbol{s}))||_1+LPIPS(R(G(\boldsymbol{s},\boldsymbol{s})),\boldsymbol{s})]\\
    +\mathrm{E}_{\boldsymbol{r}}[||R(G(\boldsymbol{r},\boldsymbol{r}))||_1+LPIPS(R(G(\boldsymbol{r},\boldsymbol{r})),\boldsymbol{r})].
\end{split}
\end{equation}

\textbf{Total Loss.} The total loss $L_D$, $L_G$ and $L_R$ for discriminator $D$, generator $G$ and rectification module $R$ can be formulated as follows:
\begin{equation}
    L_D=\lambda_{D}^{adv}L_D^{adv},
\end{equation}
\begin{equation}        
L_G=\lambda_{G}^{adv}L_G^{adv}+\lambda_{G}^{cyc}L_G^{cyc}+\lambda_{G}^{mk}L_G^{mk}+\lambda_{G}^{reg}L_G^{reg},
\end{equation}
\begin{equation} L_R=\lambda_{R}^{adv}L_R^{adv}+\lambda_{R}^{mk}L_R^{mk}+\lambda_{R}^{reg}L_R^{reg},
\end{equation}
where $\lambda's$ are hyper-parameters to balance different losses. 

\subsection{Target Model Training}
Let $\mathcal{D}_{t}=\{(\boldsymbol{x}_i,y_i)\}^N_{i=1}$ denotes the original training set containing $N$ benign samples. To poison a benign sample $(\boldsymbol{x_t}, y_t)$, we implant the trigger into the sample and change its label to the target label, resulting in the transformation:
\begin{equation}
    (\boldsymbol{x_t},y_t)\Longrightarrow (G(\boldsymbol{x_t}), \eta(y_t)),
\end{equation}
where $G(\cdot)$ represents the trigger generation function and $\eta(\cdot)$ represents the target label transformation function. We poison a portion of benign training samples, forming a poisoned dataset $\mathcal{D}_{p}$. $\mathcal{D}_{m}$ denotes the subset of $\mathcal{D}_{p}$ containing all malicious samples, and $\mathcal{D}_{b}$ denotes the remaining benign samples in $\mathcal{D}_{p}$. The poisoning rate $\gamma = |\mathcal{D}_{m}|/|\mathcal{D}_{p}|$ indicates the proportion of the poisoned samples in the dataset.

The main objective of the target model in the backdoor training process is to inject the backdoor into target models, causing them to incorrectly predict target labels for malicious samples while behaving normally on benign samples. Consequently, the training objective can be formulated as follows:
\begin{equation}
\begin{split}
    \min_{\theta}{\mathrm{E}_{(\boldsymbol{x},y)\in{\mathcal{D}_{p}}}{L^{ce}(f_{\theta}(\boldsymbol{x}),y)}},
\end{split}
\end{equation}
where $L^{ce}$ denotes the cross-entropy loss, $f_{\theta}$ represents the target model with parameters $\theta$. As evident from the above objective, only the poisoned dataset is required for training without controlling the process. However, such supervised training can only partially narrow the representation gap between the poisoned samples and the benign samples with the target label. Therefore, we utilize the target model as guidance to fine-tune the trigger generator.

\subsection{Generator Fine-tuning and Data Updating}
We introduce a perceptual loss within the pre-training framework of generator training (as mentioned in section \ref{4.2}), aiming to optimize the generation of malicious samples. The perceptual loss utilizes cosine similarity to quantify the difference in representation between the malicious samples and the benign samples with the target label. Specifically, we select benign samples with the target label from the training set as guidance samples $\boldsymbol{x_g}$. Simultaneously, we augment the malicious samples $G(\boldsymbol{s},\boldsymbol{r})$ with diverse random Gaussian noise to enhance the robustness of the generator. With both the features of guidance samples and the augmented malicious samples, we can formulate the perceptual loss as follows:
\begin{equation}
\begin{split}
    L_{G}^{per}=\mathrm{E}_{(\boldsymbol{s},\boldsymbol{r}),\boldsymbol{x_g},\boldsymbol{\psi}}[1-\cos{[M(\boldsymbol{x_g}),M(G(\boldsymbol{s},\boldsymbol{r})+\boldsymbol{\psi})}]]\\ 
    +\mathrm{E}_{(\boldsymbol{s},\boldsymbol{r}),\boldsymbol{x_g},\boldsymbol{\psi}}[1-\cos{[M(\boldsymbol{x_g}),M(G(\boldsymbol{r},\boldsymbol{s})+\boldsymbol{\psi})]}],
\end{split}
\end{equation}
where $M$ represents the feature extractor of the target model, and $\boldsymbol{\psi}$ represents the random Gaussian noise with predetermined mean and variance. 

Also, we need to constrain the feature generated by the rectification module $R$. The perceptual loss of $R$ is formulated as follows:
\begin{equation}
\begin{split}
L_{R}^{per}=\mathrm{E}_{(\boldsymbol{s},\boldsymbol{r})}[1-\cos[M(\boldsymbol{s}),M(R(G(\boldsymbol{s},\boldsymbol{r})))]]\\
    +\mathrm{E}_{(\boldsymbol{s},\boldsymbol{r})}[1-\cos[M(\boldsymbol{r}),M(R(G(\boldsymbol{r},\boldsymbol{s})))]].
\end{split}
\end{equation}

As such, the total loss function of generator $G$ and rectification module $R$ can be newly formulated as follows:
\begin{equation}
\resizebox{0.9\linewidth}{!}{$
L_G=\lambda_{G}^{adv}L_G^{adv}+\lambda_{G}^{cyc}L_G^{cyc}+\lambda_{G}^{mk}L_G^{mk}+\lambda_{G}^{reg}L_G^{reg}+\lambda_{G}^{per}L_G^{per},
$}
\end{equation}
\begin{equation} 
L_R=\lambda_{R}^{adv}L_R^{adv}+\lambda_{R}^{mk}L_R^{mk}+\lambda_{R}^{reg}L_R^{reg}+\lambda_{R}^{per}L_R^{per},
\end{equation}
where $\lambda's$ are hyper-parameters to balance different losses. 

\subsection{Adaptive Attack}
\label{section: adaptive attack}
Across a broad spectrum of poisoning-based attacks, malicious and benign samples often form distinct clusters in the feature space, a phenomenon known as feature space separability. Many existing defense mechanisms rely on the assumption of feature space separability. However, our method introduces a novel adaptive method to challenge this assumption.

Specifically, we construct a reference set comprising multiple reference images. For each original sample, we employ normalized mutual information to select the most suitable reference image. Guided by different reference images, the generated triggers also vary. By enhancing trigger diversification, the feature representations of malicious samples become more dispersed, thereby mitigating the latent separation in the feature space. 

To alleviate the side effect of trigger diversification on attack effectiveness, we opt to use the most frequently used reference image from the reference set during the inference stage. By reducing the complexity of identifying triggers, target models achieve higher attack effectiveness during the inference stage. Further details are provided in Algorithm~\ref{alg: select}.

\begin{algorithm}[ht]
    \caption{Adaptive Selection and Data Poisoning}
    \label{alg: select}
    \textbf{Input}: Clean Dataset $\mathcal{D}_{c}$, Reference Set $\mathcal{R}$, Trigger Generator $G_\psi$, Target Model $M_\theta$, Classifier $C_\phi$\\
    \textbf{Parameter}: Injection Ratio $\gamma$\\
    \textbf{Output}: Poisoned Dataset $\mathcal{D}_{p}$
    \begin{algorithmic}[1] 
        \STATE Sample subset $\mathcal{D}_{b}$ from $\mathcal{D}_{c}$.
        \FOR{$s_i  \in {\mathcal{D}_{b}}$}
        \STATE Compute the normalized mutual information (NMI) with each image in the reference set $\mathcal{R}$.
        \STATE Select the image with the highest NMI as the reference image: $r_i =\arg \max_{r_j\in{\mathcal{R}}}{NMI(s_i, r_j)}$.
        \STATE Poison the sample using generator $G$: $s_i=G(s_i,r_i)$.
        \ENDFOR
        \STATE Replace the poisoned subset $\mathcal{D}_{b}$ with the original samples in clean dataset $\mathcal{D}_{c}$ to form the poisoned dataset $\mathcal{D}_{p}$.
        \STATE \textbf{return} $\mathcal{D}_{p}$
    \end{algorithmic}
\end{algorithm}

\section{Experiments}

\subsection{Experimental Setup}
\noindent \textbf{Datasets.} In the generator training phase, we adopt the Makeup Transfer (MT) Dataset~\cite{li2018beautygan} consisting of 2,719 makeup images and 1,115 non-makeup images. In the backdoor training phase, we employ two widely-used facial datasets: PubFig~\cite{kumar2009attribute} and VGGFace2~\cite{cao2018vggface2}. PubFig is a medium-scale real-world facial dataset consisting of 58,797 images of 200 identities.  VGGFace2 is a large-scale facial dataset containing nearly 3.31 million images of 9,131 identities. Due to the imbalanced categories within the dataset, it is necessary to filter facial datasets before training. For simplicity, we choose 62 identities with the largest number and randomly select 72 high-quality images per identity from PubFig, and we choose 270 identities with the largest number and randomly select 500 high-quality images per identity from VGGFace2.

\noindent \textbf{Models.} We conduct experiments using three target models commonly employed in face recognition: Inception-v3~\cite{szegedy2016rethinking}, ResNet-50~\cite{he2016deep}, and VGG-16~\cite{simonyan2014very}.

\noindent \textbf{Baseline.} We benchmark our attack against established methods, including BadNets~\cite{gu2019badnets}, Blend~\cite{chen2017targeted}, ReFool~\cite{liu2020reflection}, SIG~\cite{barni2019new}, ISSBA~\cite{li2021invisible} and WaNet~\cite{nguyen2020wanet}. BadNets and Blend are among the two most commonly used backdoor attacks. ReFool and SIG represent prominent clean-label attacks. ISSBA and WaNet are invisible sample-specific attacks. For fair comparisons, we exclude training-controlled attacks.

\noindent \textbf{Implement Details.} In the generator training phase, we employ Adam as the optimizer with a learning rate of 0.0002 for all modules. In the backdoor training phase, we switch to SGD as the optimizer, starting with a learning rate of 0.01 and scheduling it to decrease by a factor of 0.1 every 50 epochs. For preprocessing, we perform face alignment, crop the central faces, and resize to $224\times224$. We maintain a consistent poisoning rate of $\gamma=10\%$ and designate target label $y_t=0$ for all attack experiments. A summary of MakeupAttack is given in Algorithm~\ref{alg: makeupattack}.

\begin{algorithm}[ht]
    \caption{MakeupAttack Backdoor Attack}
    \label{alg: makeupattack}
    \textbf{Input}: Generator Training Set $\mathcal{D}_{t}$, Clean Dataset $\mathcal{D}_{c}$, Reference Set $\mathcal{R}$, Trigger Generator $G_\psi$, Target Model $M_\theta$, Classifier $C_\phi$\\
    \textbf{Parameter}: Injection Ratio $\gamma$, Total Epoch Number $E$, Interception Epoch List $L$\\
    \textbf{Output}: Backdoored Target Model $M_\theta$, Fine-tuned Trigger Generator $G_\psi$
    \begin{algorithmic}[1] 
        \STATE Pre-train the trigger generator  $G_\psi$ on $\mathcal{D}_{t}$.
        \STATE Generate poisoned dataset $\mathcal{D}_{p}$ based on clean dataset $\mathcal{D}_{c}$ according to Algorithm \ref{alg: select}.
        \FOR{$i$=1,...,$E$}
        \STATE Train the target model $M_\theta$ as well as its classifier $C_\phi$ using simple cross-entropy loss.
        \IF{$i$ in $L$}
        \STATE Fine-tune the trigger generator  $G_\psi$.
        \STATE Update the poisoned dataset $\mathcal{D}_{p}$ with the fine-tuned generator $G_\psi$ according to Algorithm \ref{alg: select}.
        \ENDIF
        \ENDFOR
        \STATE \textbf{return} $M_{\theta}$,$C_\phi$, $G_\psi$
    \end{algorithmic}
\end{algorithm}

\begin{table*}[ht]
\centering
\caption{Experimental results on PubFig and VGGFace2 datasets, measuring attack success rate (ASR) and benign accuracy (BA) in percentage. \textcolor{red}{Attack failures} (ASR below $70\%$) are highlighted in red. The results of \textcolor{blue}{our method} are highlighted in blue. $\dag$ denotes the variant where the trigger generator is not fine-tuned, and malicious samples are not updated during the entire backdoor training process.}
\resizebox{0.8\textwidth}{!}{%
\begin{tabular}{c||c||cccccc||cc}
\toprule
\multirow{2}{*}{Dataset $\downarrow$} & Network $\rightarrow$ & \multicolumn{2}{c}{Inception-v3}    & \multicolumn{2}{c}{ResNet-50}  & \multicolumn{2}{c||}{VGG-16} & \multicolumn{2}{c}{Average} \\ 
                          & Attack $\downarrow$       & ASR(\%) & BA(\%)    & ASR(\%) & BA(\%) 
                          & ASR(\%) & BA(\%) & ASR(\%) & BA(\%) \\ \midrule
\multirow{10}{*}{PubFig}  & Clean Model & $-$ & 92.40 & $-$ & 89.17 & $-$ & 85.48 & $-$ & 89.02 \\
                          & BadNets 
                          & \textbf{100.00} & \textbf{92.17} 
                          & \textbf{100.00} & 83.64 
                          & \textbf{100.00} & \textbf{85.25} 
                          & \textbf{100.00} & 87.02     \\
                          & Blend 
                          & \textbf{100.00} & \underline{91.47} 
                          & \textbf{100.00} & \underline{86.18} 
                          & \textbf{100.00} & 84.79  
                          & \textbf{100.00} & \underline{87.48} \\
                          & \cellcolor{red!10} SIG  
                          & \cellcolor{red!10} 3.23 & \cellcolor{red!10} 88.94
                          &\cellcolor{red!10} 13.59 & \cellcolor{red!10} 83.64 
                          & \cellcolor{red!10} 16.36 & \cellcolor{red!10} 84.71 
                          & \cellcolor{red!10} 11.06 & \cellcolor{red!10} 85.76       \\
                          &\cellcolor{red!10} Refool        
                          &\cellcolor{red!10} 17.28 &\cellcolor{red!10} 91.47 
                          &\cellcolor{red!10} 25.88 &\cellcolor{red!10} 84.79 
                          &\cellcolor{red!10} 31.80 &\cellcolor{red!10} 79.95 
                          &\cellcolor{red!10} 24.99 &\cellcolor{red!10} 85.40 \\
                          &\cellcolor{red!10} WaNet        
                          &\cellcolor{red!10} 19.59   & \cellcolor{red!10} 84.79
                          &\cellcolor{red!10} 23.96 & \cellcolor{red!10} 79.49
                          &\cellcolor{red!10} 27.19 & \cellcolor{red!10} 77.88 
                          &\cellcolor{red!10} 23.58 & \cellcolor{red!10} 80.72\\
                          &\cellcolor{red!10} ISSBA 
                          &\cellcolor{red!10} 63.82 &\cellcolor{red!10} 66.82 
                          & \underline{99.31} & 73.04 
                          &\cellcolor{red!10} 11.06 &\cellcolor{red!10} 67.74
                          &\cellcolor{red!10} 58.06 &\cellcolor{red!10} 69.20\\
                          &\cellcolor{blue!10}MakeupAttack\dag 
                          &\cellcolor{blue!10} 97.00 &\cellcolor{blue!10} 90.32
                          &\cellcolor{blue!10} 97.31 &\cellcolor{blue!10} 85.24 
                          &\cellcolor{blue!10} 91.94 &\cellcolor{blue!10} 79.72
                          &\cellcolor{blue!10} 95.41 & \cellcolor{blue!10} 85.09 \\
                          &\cellcolor{blue!10}MakeupAttack  
                          &\cellcolor{blue!10} \underline{97.47} &\cellcolor{blue!10} \textbf{92.17} 
                          &\cellcolor{blue!10} 98.16  &\cellcolor{blue!10} \textbf{90.74}
                          &\cellcolor{blue!10} \underline{92.47} &\cellcolor{blue!10} \textbf{85.25}
                          &\cellcolor{blue!10} \underline{96.03} &\cellcolor{blue!10} \textbf{89.39}
                          \\ \midrule
\multirow{10}{*}{VGGFace2}
                          & Clean Model   
                          & - & 98.45                      
                          & - & 98.52                
                          & - & 99.16
                          & - & 98.71\\
                          & BadNets 
                          & 99.50 & \underline{97.79}    
                          & 99.51 & 98.35
                          & 99.68 & 98.90
                          & 99.56 & 98.34\\
                          & Blend 
                          & \textbf{100.00} & \textbf{97.96}
                          & \textbf{100.00} & \underline{98.42}  
                          & \textbf{100.00} & 98.92
                          & \textbf{100.00} & \textbf{98.43}\\
                          &\cellcolor{red!10} SIG 
                          &\cellcolor{red!10}15.61 &\cellcolor{red!10} 97.72
                          &\cellcolor{red!10}31.51 &\cellcolor{red!10} 98.24
                          & \textbf{100.00} & 98.93
                          &\cellcolor{red!10} 49.04 &\cellcolor{red!10} 98.30\\
                          &\cellcolor{red!10} Refool
                          &\cellcolor{red!10} 46.10 &\cellcolor{red!10} 97.65
                          &\cellcolor{red!10} 58.79 &\cellcolor{red!10} 98.26
                          & 99.35 & 98.90
                          &\cellcolor{red!10} 68.08 &\cellcolor{red!10} 98.27 \\
                          & WaNet         
                          & 99.66 & 97.55
                          & \textbf{100.00} & 98.39
                          & \textbf{100.00} & \textbf{99.10}
                          & \underline{99.88} & 98.34\\
                          & ISSBA 
                          & \textbf{100.00} & 80.80
                          & \textbf{100.00} & 73.24 
                          & \textbf{100.00} & 76.62
                          & \textbf{100.00} & 76.89\\
                          &\cellcolor{blue!10} MakeupAttack\dag 
                          &\cellcolor{blue!10} 99.56 &\cellcolor{blue!10} 97.34
                          &\cellcolor{blue!10} 99.70 &\cellcolor{blue!10} 98.12
                          &\cellcolor{blue!10} 99.75 &\cellcolor{blue!10} 98.81
                          &\cellcolor{blue!10} 99.67 &\cellcolor{blue!10} 98.09\\
                          &\cellcolor{blue!10} MakeupAttack
                          &\cellcolor{blue!10} \underline{99.70} &\cellcolor{blue!10} 97.66
                          &\cellcolor{blue!10} \underline{99.89} &\cellcolor{blue!10} \textbf{ 98.47}
                          &\cellcolor{blue!10} \underline{99.90} &\cellcolor{blue!10} \underline{98.94}
                          &\cellcolor{blue!10} 99.83 & \cellcolor{blue!10} \underline{98.35}\\ \bottomrule
\end{tabular}%
}
\label{tab: effectiveness}
\end{table*}

\subsection{Attack Experiments} 
We evaluate attack effectiveness with the attack success rate (ASR) and benign accuracy (BA). ASR indicates the ratio of malicious samples incorrectly predicted as the target label, while BA indicates the ratio of benign samples correctly predicted. As shown in Table~\ref{tab: effectiveness}, our method successfully attacks various target models across multiple datasets, showcasing its effectiveness. The average ASR of MakeupAttack reaches 98\%, sufficient to implant backdoors into target models. With sufficient training data, ASR can surpass 99.7\%, even exceeding typical pixel space attacks. Moreover, the difference in BA between clean models and those attacked by MakeupAttack ranges from -0.82 to +1.57, minimally impacting model performance on benign samples. 

Due to the characteristics of facial datasets, clean-label attacks like Refool and SIG are ineffective on face recognition models. Additionally, due to the insufficient samples in datasets, the advanced sample-specific attacks ISSBA and WaNet fail to guarantee the attack robustness. Additionally, ISSBA generally leads to compromised performance on benign samples. In contrast, our method demonstrates robustness across different datasets and network structures. Although BadNets and Blend exhibit strong attack effectiveness, their triggers are conspicuous and easily detectable. On the contrary, MakeupAttack prioritizes naturalness and stealthiness, remaining imperceptible to detection systems.

Furthermore, experimental results highlight the significant impact of generator fine-tuning and data updating on attack effectiveness. Through iterative training, our method improves ASR by 0.14-0.85 and BA by 0.13-1.85, achieving nearly optimal BA alongside high ASR. These results underscore that our method facilitates learning on benign samples, thus maintaining excellent performance on BA.

\subsection{Defense Experiments}
We test the resistance capabilities of MakeupAttack against commonly used defense methods, including STRIP~\cite{gao2019strip}, Signature Spectral~\cite{tran2018spectral}, Fine-Pruning~\cite{liu2018fine} and CLP~\cite{zheng2022data}.

\begin{figure}[ht]
    \centering
    \includegraphics[width=1\columnwidth]{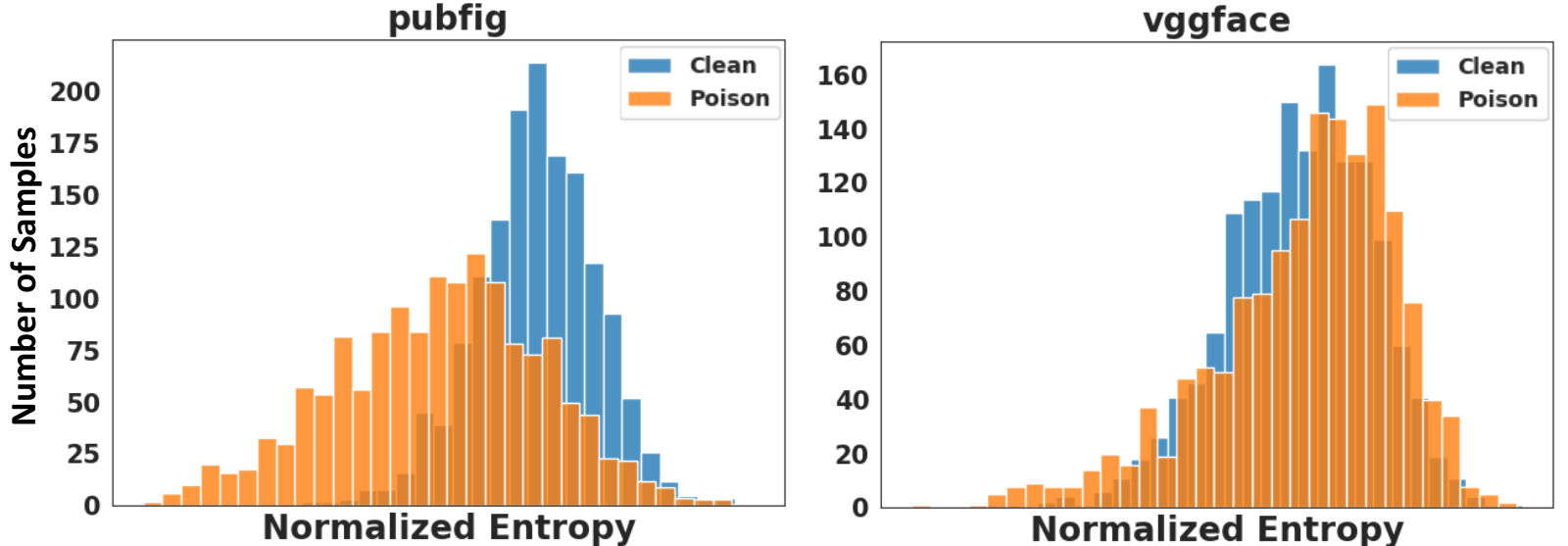}
    \caption{Experimental results of STRIP.}
    \label{fig: strip}
\end{figure}

\noindent\textbf{Resistance to STRIP.} STRIP assumes that the predictions made by a backdoored model exhibit stability on malicious samples. It detects such samples by computing the entropy of classification probabilities after overlaying random samples. Figure~\ref{fig: strip} illustrates that STRIP fails to establish a threshold to distinguish between benign and malicious samples, enabling our attack to bypass the detection successfully. 

\begin{figure}[htbp]
    \centering
    \includegraphics[width=1\columnwidth]{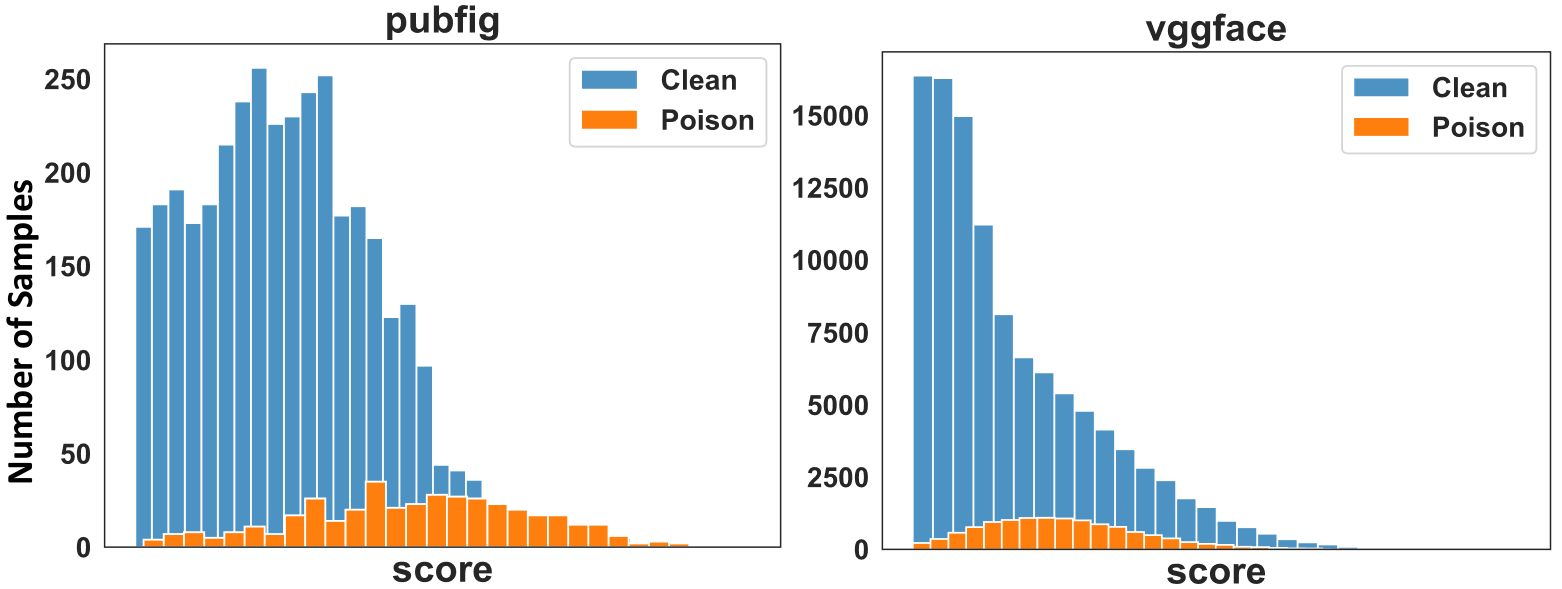}
    \caption{Experimental results of Signature Spectral.}
    \label{fig: spectral}
\end{figure}

\noindent\textbf{Resistance to Signature Spectral.} Signature Spectral detects malicious samples by identifying detectable traces in the spectrum of the covariance of feature representations. By computing the correlation of features and deriving the top singular value as the outlier score for each sample, the method assesses the likelihood of a sample being malicious. As depicted in Figure~\ref{fig: spectral}, malicious and benign samples are mixed in the outlier score distribution, rendering the setting of an appropriate threshold unfeasible for distinguishing between the two. 

\noindent \textbf{Resistance to SentiNet.} SentiNet~\cite{chou2020sentinet} identifies triggers based on the similarity of Grad-Cam of various malicious samples poisoned by the same attack. Figure \ref{fig: grad} demonstrates that Grad-CAM can successfully distinguish trigger regions of BadNets and Blend but fails to detect the trigger of our attack. Additionally, the visualization shows that the face recognition model attacked by our method can pay more attention to crucial facial areas rather than trigger regions.

\noindent\textbf{Resistance to Fine-pruning.} Fine-pruning identifies compromised neurons by analyzing the abnormality of activation values and mitigates the backdoor by pruning these neurons without decreasing benign accuracy. As depicted in Figure \ref{fig: fine-pruning}, Fine-pruning is unable to eliminate the backdoor injected by MakeupAttack without sacrificing performance on benign samples.

\noindent \textbf{Resistance to CLP.} CLP detects potential backdoor channels in a data-free manner and repairs attacked models via simple channel pruning. Table~\ref{tab: CLP} demonstrates that CLP mitigates the attack capabilities of MakeupAttack while significantly compromising model performance on benign samples, effectively resisting CLP.

\begin{figure}[ht]
    \centering
    \includegraphics[width=0.9\columnwidth]{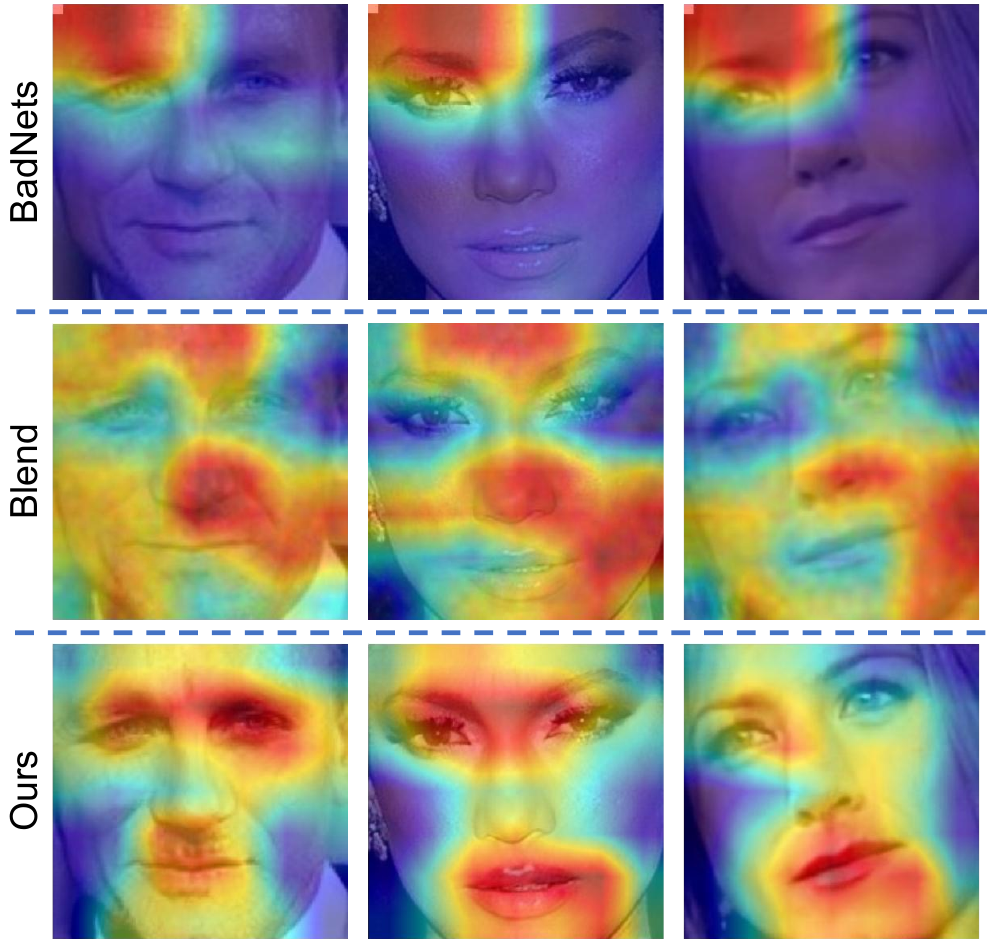}
    \caption{The attention maps of various poisoned samples.}
    \label{fig: grad}
\end{figure}

\begin{figure}[ht]
    \centering
    \includegraphics[width=0.9\columnwidth]{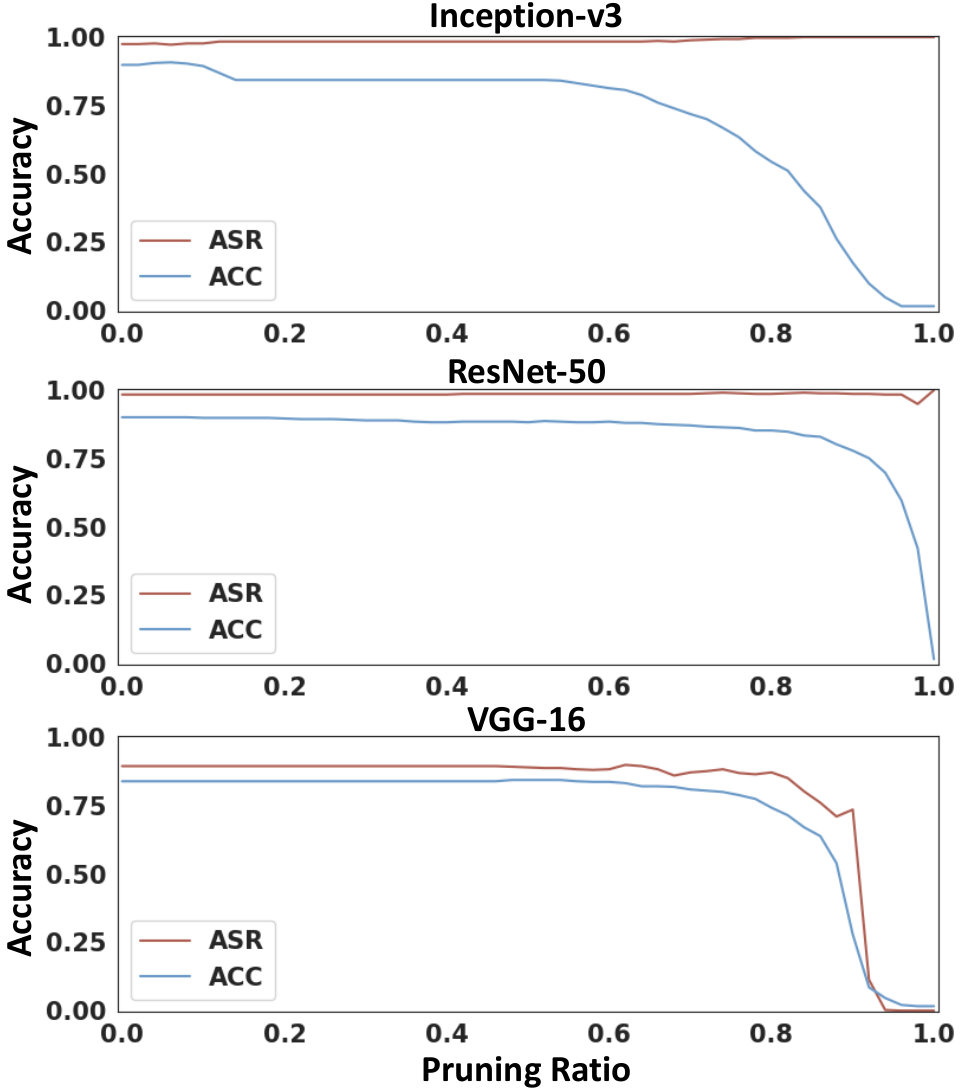}
    \caption{Experimental results of Fine-pruning.}
    \label{fig: fine-pruning}
\end{figure}

\begin{table}[ht]
\centering
\caption{Experimental results of CLP.}
\resizebox{0.85\columnwidth}{!}{%
\begin{tabular}{c|cc|cc}
\toprule
\multirow{2}{*}{Dataset} & \multicolumn{2}{c|}{Backdoored} & \multicolumn{2}{c}{CLP Pruned} \\
                         & ASR(\%)            & BA(\%)             & ASR(\%)       & BA(\%)        \\ \midrule
PubFig                   & 98.16          & 90.74          & 0          & 0.04      \\ 
VGGFace2                 & 99.70          & 97.63          & 0          & 0          \\ \bottomrule
\end{tabular}%
}
\label{tab: CLP}
\end{table}

\subsection{Ablation Study}
\subsubsection{Rectification Module}
The rectification module offers certain advantages in improving attack effectiveness. As depicted in Table~\ref{tab: rectification}, employing the rectification module during the generator training phase leads to higher ASR and BA for target models, indicating a subtle yet discernible improvement in attack effectiveness.

Additionally, the utilization of the rectification modules results in more natural-looking generated images. Detailed examples are provided in Appendix~\ref{sec: rect}.

\subsubsection{Selection Mode} 
We adopt various modes for selecting reference images during the backdoor training phase, including RAND for random selection, SSIM for selection based on structure similarity index measure, and NMI for selection based on normalized mutual information. Table \ref{tab: selection} indicates that NMI is a better indicator for selection.


\begin{table}[ht]
\centering
\caption{Rectification module R and attack effectiveness.}
\resizebox{0.9\columnwidth}{!}{%
\begin{tabular}{c|c|cc}
\toprule
Dataset                   & Rectification Module R & ASR(\%) & BA(\%) \\ \midrule
\multirow{2}{*}{PubFig}   & w/o R                  &  96.08   & 84.79   \\
                          & w/ R                   & \textbf{97.31}     & \textbf{85.24}   \\ \midrule
\multirow{2}{*}{VGGFace2} & w/o R               &  99.65   &  98.05  \\
                          & w/ R                   & \textbf{99.70}    & \textbf{98.12}   \\ \bottomrule
\end{tabular}%
}
\label{tab: rectification}
\end{table}

\begin{table}[h]
\centering
\caption{Selection mode and attack effectiveness.}
\resizebox{0.8\columnwidth}{!}{%
\begin{tabular}{c|c|cc}
\toprule
Dataset                   & Selection Mode & ASR(\%) & BA(\%) \\ \midrule
\multirow{3}{*}{PubFig}   & RAND   & 96.29    & 82.72   \\
                          & SSIM           & 92.63    & 81.57   \\
                          & NMI            &  \textbf{97.31}    & \textbf{85.24}     \\ \midrule
\multirow{3}{*}{VGGFace2} & RAND           & 99.55    & 97.97   \\
                          & SSIM           & 99.25    & 98.05   \\
                          & NMI            & \textbf{99.70}    & \textbf{98.12}   \\ \bottomrule
\end{tabular}%
}
\label{tab: selection}
\end{table}

\section{Dataset Release}

For reproducing and further developing our method, we have constructed two high-quality malicious datasets. Firstly, we select high-quality facial images from PubFig and VGGFace2, covering various lighting conditions, backgrounds, poses, and expressions. Subsequently, we employ our proposed framework to poison and update these raw images (following Algorithm~\ref{alg: select}). Finally, we use the integrated facial processing tool InsightFace~\cite{guo2018stacked} to align faces and compile the images into two malicious datasets. 

Given the universal applicability of our transfer method for both male and female faces (additional transferred examples are available in Appendix~\ref{sec: vis}), concerns regarding conspicuousness on male faces are alleviated.


\section{Conclusion}
In this paper, we propose MakeupAttack, a novel feature space backdoor attack designed for face recognition models. Our approach leverages makeup transfer to craft natural triggers, enabling subtle manipulation of feature representation. To capture subtle trigger patterns, we introduce an iterative training paradigm tailored for black-box attack scenarios. Additionally, we employ an adaptive selection method to enhance trigger diversity, facilitating evasion of various defense mechanisms. Extensive experiments and visualizations validate the effectiveness, robustness, naturalness, stealthiness, and defense resistance of our method.

\clearpage
\begin{ack}
This work is supported in part by the National Natural Science Foundation of China Under Grants No. 62176253.
\end{ack}

\bibliography{ecai}

\clearpage
\section{Appendix}

\subsection{Nomenclature}
\begin{table}[ht]   
\begin{framed}
\nomenclature[D]{$\boldsymbol{s}$}{source image}
\nomenclature[D]{$\boldsymbol{r}$}{reference image}
\nomenclature[D]{$\mathcal{R}$}{reference set}
\nomenclature[D]{$\mathcal{D}_t$}{generator training set}
\nomenclature[D]{$\mathcal{D}_c$}{clean training set}
\nomenclature[D]{$\mathcal{D}_p$}{poisoned training set}
\nomenclature[D]{$\mathcal{D}_b$}{selected subset to modify}
\nomenclature[D]{$\mathcal{D}_m$}{modified malicious subset}
\nomenclature[M]{$G_{\psi}$}{trigger generator}
\nomenclature[M]{$R$}{rectification module}
\nomenclature[M]{$D_S$}{source domain discriminator}
\nomenclature[M]{$D_R$}{reference domain discriminator}
\nomenclature[M]{$M_\theta$}{target model}
\nomenclature[M]{$C_\phi$}{classifier}
\nomenclature[L]{$L^{adv}$}{adversarial loss}
\nomenclature[L]{$L^{cyc}$}{cycle consistency loss}
\nomenclature[L]{$L^{mk}$}{makeup loss}
\nomenclature[L]{$L^{reg}$}{regularization loss}
\nomenclature[L]{$L^{per}$}{perceptual loss}
\nomenclature[L]{$L^{ce}$}{cross-entropy loss}
\nomenclature[P]{$\gamma$}{poisoning rate}
\nomenclature[P]{$L$}{interception epoch list}
\nomenclature[P]{$E$}{total training epoch}
\printnomenclature
\end{framed}
\end{table}

\subsection{Experiment Configurations}
\subsubsection{Attack Configurations}
BadNets uses a static $8\times8$ white square as the trigger. Blend employs Gaussian noise with $20\%$ opacity as the trigger pattern. We follow their original attack settings for SIG, Refool, and WaNet. For ISSBA, the size of the generated trigger is set to $224\times224$, following other original settings. For MakeupAttack, the reference set comprises 16 images with different makeup styles.

During the trigger generator training phase, we adopt the same settings as the PSGAN framework and train the generator for 5 epochs on the Makeup Transfer (MT) Dataset. Adam with $\beta_1=0.5$ and $\beta_1=0.999$ is used for optimization, with a learning rate of $2\times10^{-4}$ for the generator, discriminator, and rectification module. To fine-tune the trigger generator, we randomly select 3 clean images with the target label from the dataset as guidance samples and augment them with Gaussian noise of the same distribution, twice for each fake sample generated by the trigger generator.

During the backdoor training phase, we adopt the standard training pipeline on PubFig and VGGFace2. We use SGD with a momentum of 0.9 and a weight decay of $5\times10^{-4}$ for optimization. The initial learning rate is set to 0.1, decreasing by a factor of 0.1 every 50 epochs. Backdoor training is conducted for 300 epochs. In MakeupAttack, we first train the target model with the training data poisoned by the pre-trained generator for 150 epochs. Subsequently, we fine-tune the trigger generator using the semi-trained target model and continue training for the remaining 150 epochs with the training data updated by the fine-tuned generator.  In addition, we incorporate two widely-used data augmentation methods for training, i.e. \texttt{RandomHorizontalFlip} and \texttt{RandomCrop}.

\subsubsection{Defense Configurations}

\noindent \textbf{STRIP.} We randomly select 1,500 malicious and benign samples from the original training set and another 2,000 benign samples as the auxiliary set. For each original sample, we superimpose random samples from the auxiliary set to generate 100 samples for calculating the entropy of the classification probability. We evaluate the defense efficiency of STRIP against ResNet-50 on PubFig and VGGFace2.

\noindent \textbf{Signature Spectral.} We randomly poison $10\%$ of the original training samples and calculate the outlier score of each sample. We evaluate the defense efficiency of Signature Spectral against ResNet-50 on PubFig and VGGFace2.

\noindent \textbf{SentiNet.} We utilize Grad-Cam to generate attention maps of ResNet-50 in layer4.  In the main paper, we demonstrate its resistance to SentiNet using images from PubFig. Appendix~\ref{sec: 8.3.1} includes additional verification using images from VGGFace2.

\noindent \textbf{Fine-pruning.} Block8 of Inception-v3, layer4 of ResNet-50, and layer5 of VGG-16 are chosen for pruning. We evaluate the defense efficiency of Fine-pruning on PubFig. 

\noindent \textbf{CLP.} The threshold set $\theta^{(l)}$ for the $l^{th}$ layer during channel pruning can be formulated as follows:
\begin{equation}
    \theta^{(l)}=\mu^{(l)}+u \cdot s^{(l)},
\end{equation}
where $\mu^{(l)}$ and $s^{(l)}$ are the mean and the standard deviation of the upper bound of the Channel Lipschitz Constant (UCLC) for the $l^{th}$ layer and $u$ is the only hyper-parameter $u$ set to the default value of 3. We evaluate the model performance after pruning the backdoored ResNet-50 models trained on PubFig and VGGFace2.

\noindent \textbf{Neural Cleanse.} As the target label is set to 0, we reverse triggers for label 0 on PubFig and VGGFace2 using Adam optimizer with a learning rate of 0.005. The reverse engineering process takes 100 epochs.

\subsection{More Defense Experiments}

\subsubsection{Further Analysis on SentiNet.}
\label{sec: 8.3.1}
We extend our investigation of SentiNet by examining images from the VGGFace2 dataset. The results indicate that SentiNet predominantly focuses on the trigger regions of BadNets and Blend. However, when faced with images poisoned by MakeupAttack, SentiNet primarily attends to the facial area, failing to detect the hidden trigger region of our attack.

\subsubsection{Resistance to Neural Cleanse.}
Neural Cleanse (NC)~\cite{wang2019neural} is a trigger-synthesis-based backdoor defense that attempts to reverse triggers by optimizing the pixel space. NC assumes that the reversed trigger has an abnormally small norm and is more likely to correspond to a poisoned target label. In Table~\ref{tab: nc}, we observe that triggers reversed from models poisoned by BadNets have relatively small $l_1$-norms ($<40$), whereas triggers reversed from models poisoned by our method have larger $l_1$-norm ($>160$). This indicates that while NC predominantly optimizes for facial features, it fails to capture the trigger pattern of our attack, allowing our method to bypass NC.

\begin{figure}[htbp]
    \centering
    \includegraphics[width=1\columnwidth]{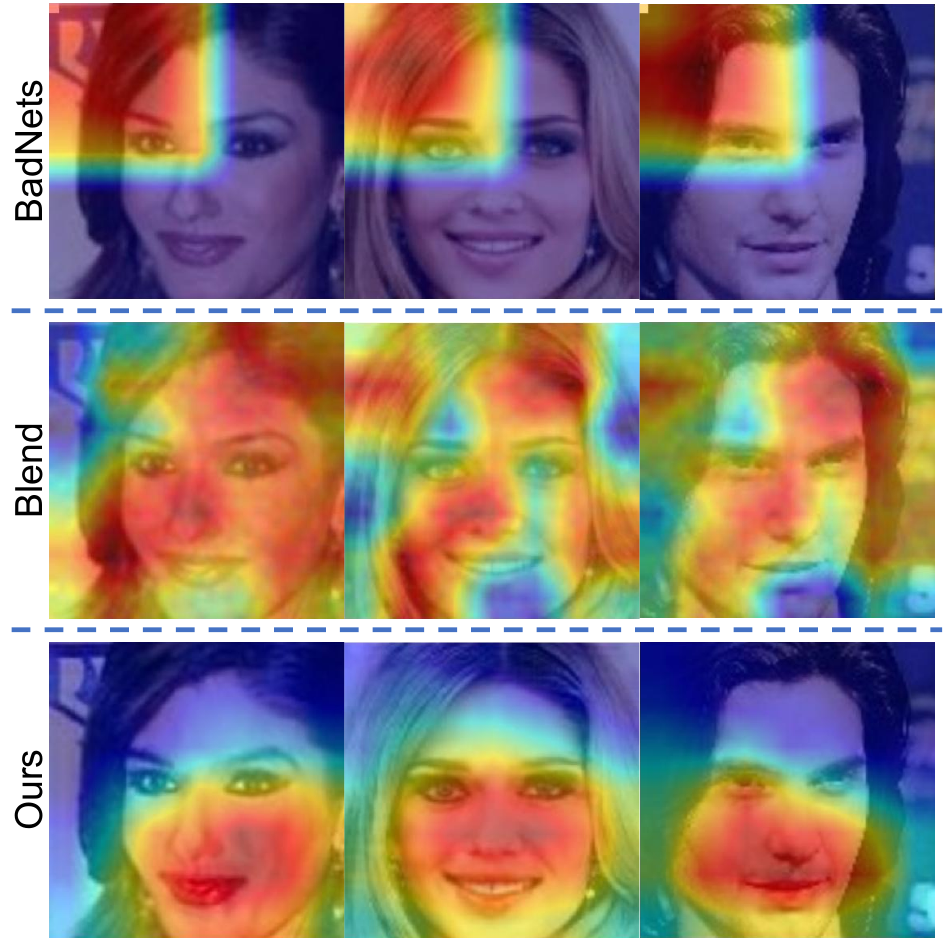}
    \caption{The attention maps of various poisoned samples from VGGFace2 generated by Grad-Cam.}
    \label{fig: grad}
\end{figure}

\begin{table}[ht]
\centering
\caption{Reversed triggers of Neural Cleanse. The numbers below each image represent the l$_1$-norm of the mask of each reversed trigger.}
\resizebox{\columnwidth}{!}{%
\begin{tabular}{ccc}
\toprule
Attack                   & PubFig & VGGFace2 \\ \midrule
\multirow{2}{*}[9ex]{Clean}   & \includegraphics[width=0.4\columnwidth,height=0.4\linewidth]{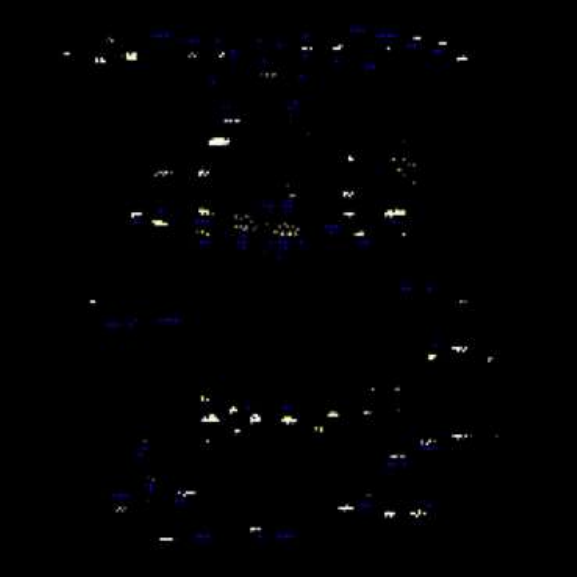} &
\includegraphics[width=0.4\columnwidth,height=0.4\linewidth]{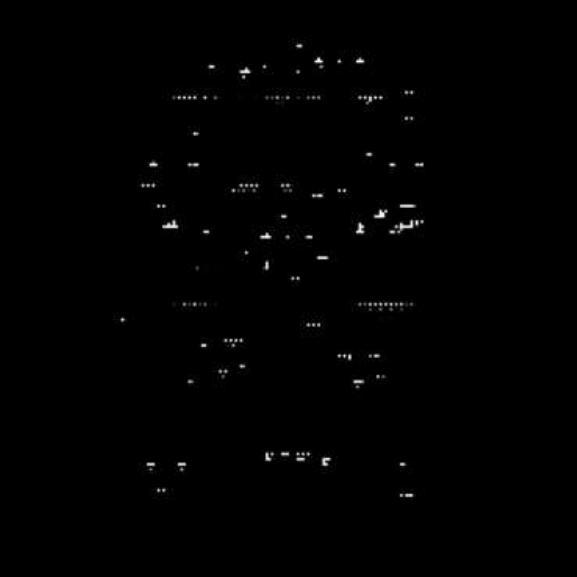}         \\
                         & l$_1$-norm: 469.82 & l$_1$-norm: 329.45   \\ \midrule
\multirow{2}{*}[9ex]{BadNets} & \includegraphics[width=0.4\columnwidth,height=0.4\linewidth]{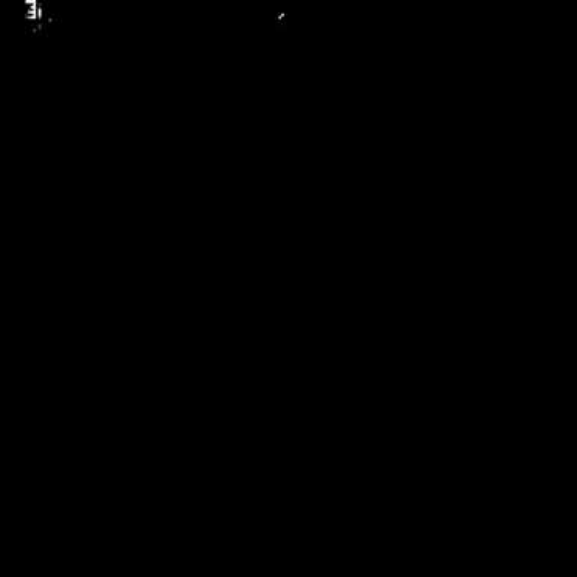}       & \includegraphics[width=0.4\columnwidth,height=0.4\linewidth]{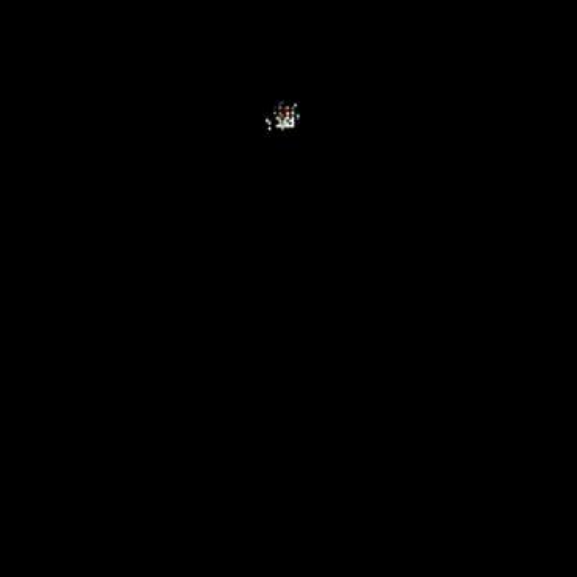}         \\
                         & l$_1$-norm: 21.44  & l$_1$-norm: 36.78    \\ \midrule
\multirow{2}{*}[9ex]{Ours}    & \includegraphics[width=0.4\columnwidth,height=0.4\linewidth]{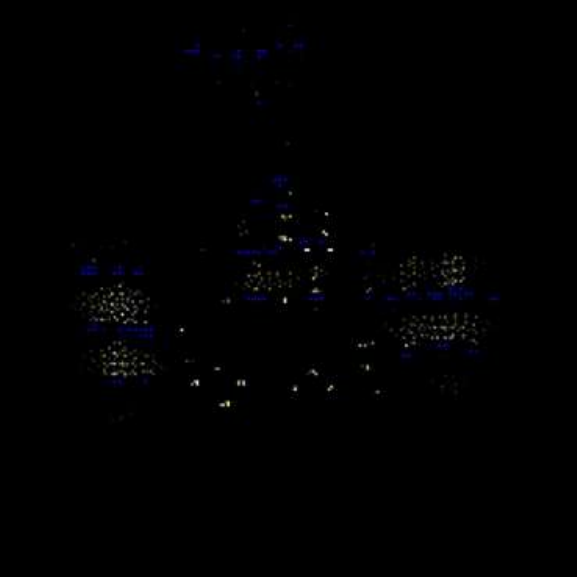}       & \includegraphics[width=0.4\columnwidth,height=0.4\linewidth]{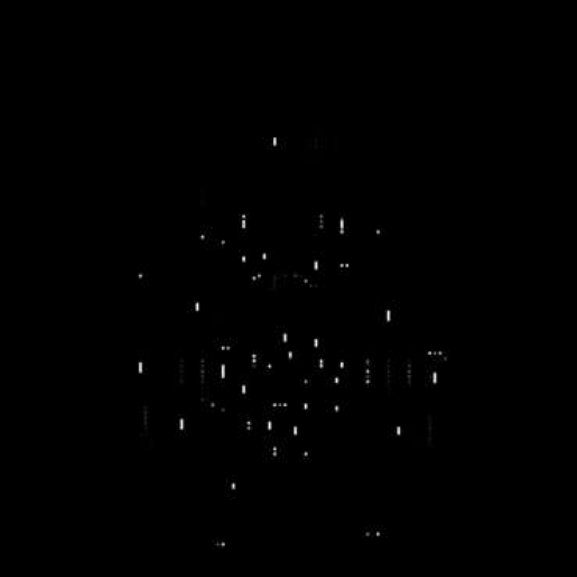}         \\
                         & l$_1$-norm: 292.33 & l$_1$-norm: 197.48   \\ \bottomrule
\end{tabular}%
}
\label{tab: nc}
\end{table}

\begin{figure}[ht]
    \centering
    \includegraphics[width=0.8\columnwidth]{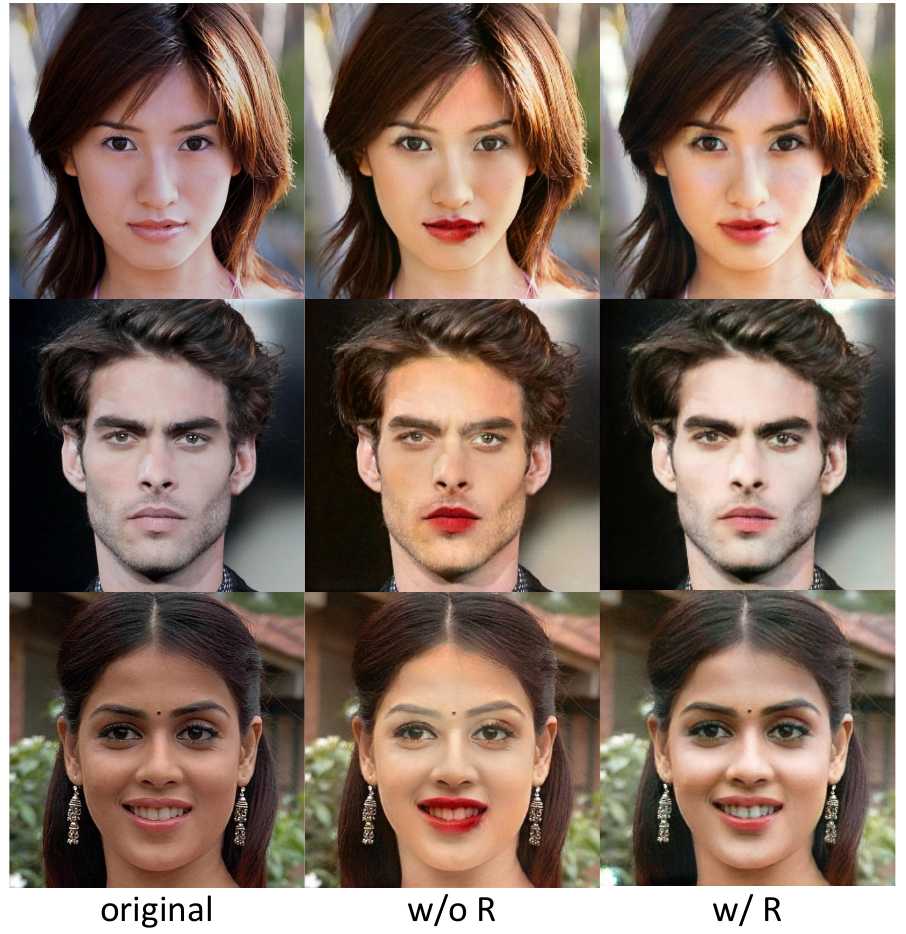}
    \caption{Ablation study for rectification module. Images generated by the generator trained with the rectification look more natural.}
    \label{fig: rec}
\end{figure}

\begin{figure}[ht]
    \centering
    \includegraphics[width=0.9\columnwidth]{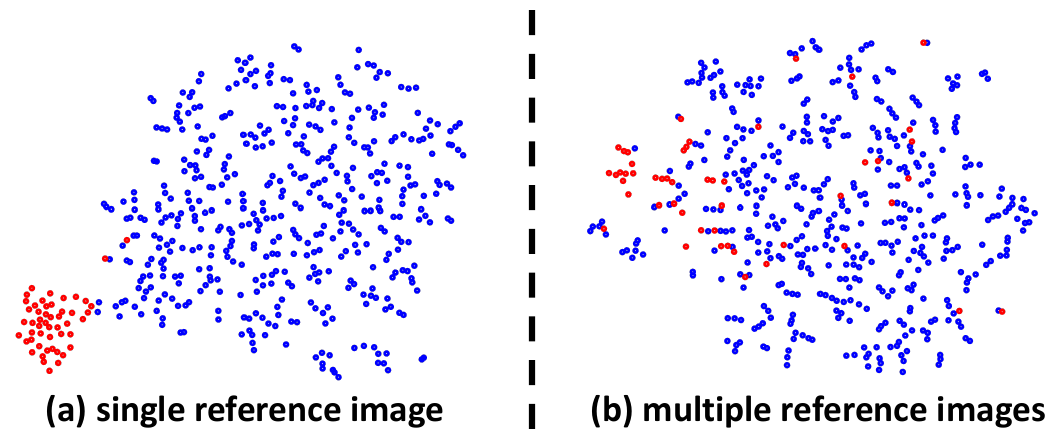}
    \caption{T-SNE visualizations of feature space separability characteristic on VGGFace2. To highlight the separation, all poison samples are denoted by red points, while blue points denote clean samples.}
    \label{fig: adapt}
\end{figure}

\begin{figure*}[ht]
    \centering
    \includegraphics[width=1.8\columnwidth]{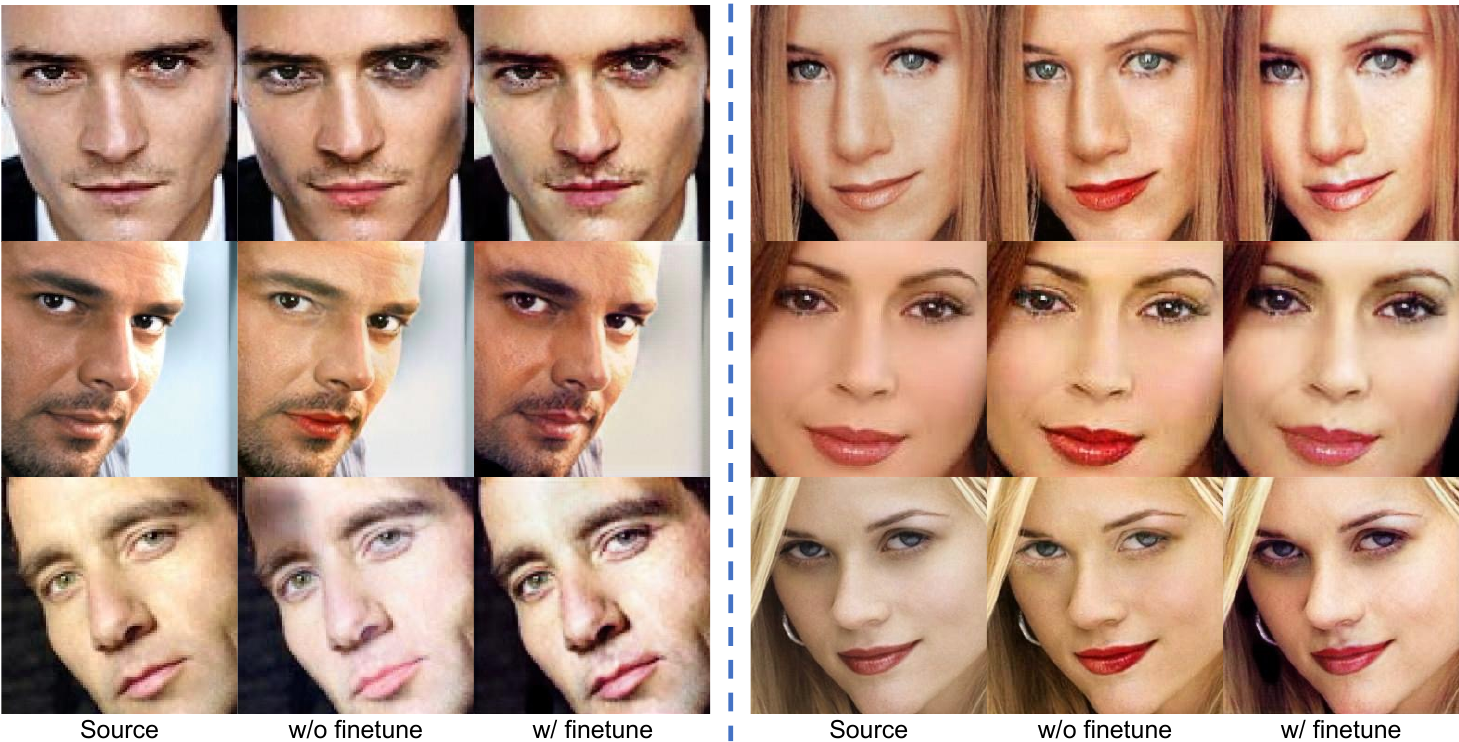}
    \caption{Additional images containing different poses, genders, and expressions for demonstrations.}
    \label{fig: vis}
\end{figure*}

\subsection{More Ablation Study Results}

\subsubsection{Rectification Module}
\label{sec: rect}
In addition to maintaining attack effectiveness, the rectification module enhances the naturalness and stealthiness of the generated samples. High-quality and high-resolution images ($1024\times1024$)  from CelebA-HQ~\cite{karras2017progressive}are selected for this analysis. Utilizing the same reference image, the benefits of the module become evident. As depicted in Figure~\ref{fig: rec}, images generated with the rectification module exhibit improved stealthiness and naturalness, closely resembling the original images.

Moreover, the makeup triggers generated with the rectification module are suitable for both genders and do not appear conspicuous on male faces, facilitating the construction of gender-uniform makeup-poisoned datasets.

\subsection{Adaptive Selection}
We compare the feature distribution between using multiple reference images and a single reference image. Figure~\ref{fig: adapt} presents the T-SNE visualization of these two modes on VGGFace2. The results illustrate that employing multiple reference images enhances trigger diversity, dispersing the feature representation of malicious samples. This divergence challenges the assumption of feature space separability for many backdoor defenses.

\subsection{More Explanations}
\subsubsection{NMI}
NMI evaluates image similarity by calculating mutual information between images, capturing the shared information and complicated relationships. It exhibits robustness to complex transformations such as nonlinear transformations and noise, making it suitable for facial datasets with uneven image quality. Therefore, NMI is widely used in scenarios requiring global information measurement, such as image fusion. On the other hand, PSNR and SSIM are better for evaluating image quality. Considering the complex facial characteristics, NMI is more suitable for selecting similar faces in our adaptive reference selection.

\subsubsection{Rectification Module and Cycle Consistency Loss}
The cycle consistency loss is critical in image-to-image translation tasks, as it facilitates learning the bidirectional mapping between the source and reference domains without supervised data. In our framework, the transferred samples exhibit both poisoning attributes and domain-specific characteristics. However, the poisoning attribute disrupts the bijection relationship between the two domains. To address this, we employ the rectification module to discard the poisoning attribute, ensuring effective cycle consistency loss. This module guarantees a complete cycle reconstruction path, retaining the semantic information of the source images. Additionally, it effectively separates the poisoning attribute from the domain-specific characteristics. This separation allows the poisoning attribute to be better embedded into malicious samples, making the poisoned trigger easier to learn by target models and thus improving attack effectiveness.

\subsection{More Visualization Results.}
\label{sec: vis}
More visualization samples are available to demonstrate the stealthiness and naturalness of our method. As shown in Figure \ref{fig: vis}, our method exhibits robustness to pose and expression variations. Moreover, our method addresses the gender bias issue common in makeup transfer tasks, where images of females often exhibit better visual quality due to imbalances in the makeup transfer training dataset. Additionally, our method can effectively transfer makeup even when the source image already contains makeup, maintaining naturalness.

We conduct a comparison of generation performance before and after fine-tuning the generator. Results indicate that the generator before fine-tuning yields more obvious makeup effects, while the fine-tuned generator achieves better naturalness and stealthiness. Samples generated by the fine-tuned generator closely resemble the original images, mitigating artifacts in the samples before fine-tuning.

\end{document}